\documentclass[10pt,twocolumn,letterpaper]{article}

\usepackage{btas}
\usepackage{times}
\usepackage{epsfig}
\usepackage{graphicx}
\usepackage{amsmath}
\usepackage{amssymb}

\usepackage{epstopdf}
\usepackage[pagebackref=true,breaklinks=true,letterpaper=true,colorlinks,bookmarks=false]{hyperref}

\btasfinalcopy 

\newcommand\numkeypoints{115,000}
\newcommand\numtest{77,228}
\newcommand\numtestid{2,133}
\newcommand\umdnumim{367,888}
\newcommand\umdnumid{8,277}
\newcommand\umremovedim{12,789}
\newcommand\umremovedsub{156}

\ifbtasfinal\fi
\begin{document}

\title{UMDFaces: An Annotated Face Dataset for Training Deep Networks}

\author{Ankan Bansal \hspace{0.5cm} Anirudh Nanduri \hspace{0.5cm} Carlos D. Castillo \hspace{0.5cm} Rajeev Ranjan \hspace{0.5cm} Rama Chellappa \\
	University of Maryland, College Park\\
	{\tt\small \{ankan,snanduri,carlos,rranjan1,rama\}@umiacs.umd.edu}
}

\maketitle
\thispagestyle{empty}

\begin{abstract}
	Recent progress in face detection (including keypoint detection), and recognition is mainly being driven by (i) deeper convolutional neural network architectures, and (ii) larger datasets. However, most of the large datasets are maintained by private companies and are not publicly available. The academic computer vision community needs larger and more varied datasets to make further progress. 
	
	In this paper we introduce a new face dataset, called UMDFaces, which has \umdnumim ~annotated faces of \umdnumid ~subjects. We also introduce a new face recognition evaluation protocol which will help advance the state-of-the-art in this area. We discuss how a large dataset can be collected and annotated using human annotators and deep networks. We provide human curated bounding boxes for faces. We also provide estimated pose (roll, pitch and yaw), locations of twenty-one key-points and gender information generated by a pre-trained neural network. In addition, the quality of keypoint annotations has been verified by humans for about \numkeypoints~images.  Finally, we compare the quality of the dataset with other publicly available face datasets at similar scales.
\end{abstract}

\section{Introduction}
Current deep convolutional neural networks are very high capacity representation models and contain millions of parameters. Deep convolutional networks are achieving state-of-the-art performance on many computer vision problems \cite{lfwsurvey, resnet, stochastic}. These models are extremely data hungry and their success is being driven by the availability of large amounts of data for training and evaluation. The ImageNet dataset \cite{imagenet} was among the first large scale datasets for general object classification and since it's release has been expanded to include thousands of categories and millions of images. Similar datasets have been released for scene understanding \cite{lsun, places}, semantic segmentation \cite{pascal-voc-2012, mscoco}, and object detection \cite{pascal-voc-2012, imagenet, kitti}. 

Recent progress in face detection, and recognition problems is also being driven by deep convolutional neural networks and large datasets \cite{lfwsurvey}. However, the availability of the largest datasets and models is restricted to corporations like Facebook and Google. Recently, Facebook used a dataset of about 500 million images over 10 million identities for face identification \cite{web-scale}. They had earlier used about 4.4 million images over 4000 identities for training deep networks for face identification \cite{deepface}. Google also used over 200 million images and 8 million identities for training a deep network with 140 million parameters \cite{facenet}. But, these corporations have not released their datasets publicly.

\begin{figure*}
	\begin{center}
		\includegraphics[width=0.6\linewidth]{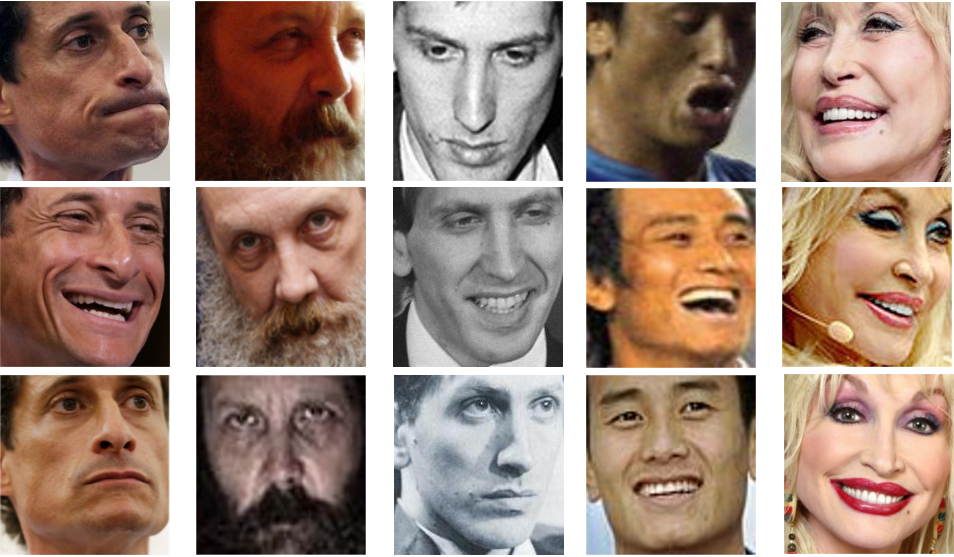}
	\end{center}
	\caption{Few samples from the dataset discussed in the paper. Each column represents variations in pose and expression of images of a subject.}
	\label{fig:five_people}
\end{figure*}

The academic community is at a disadvantage in advancing the state-of-the-art in facial recognition problems due to the unavailability of large high quality training datasets and benchmarks. Several groups have made significant contributions to overcome this problem by releasing large and diverse datasets. Sun \etal released the CelebFaces+ dataset containing a little over 200,000 images of about 10,000 identities \cite{celebfaces+}. In 2014 Dong \etal published the CASIA WebFace database for face recognition which has about 500,000 images of about 10,500 people \cite{casia}. Megaface 2 \cite{megaface2} is a recent large dataset which contains 672,057 identities with about 4.7 million images. YouTube Faces \cite{ytf} is another dataset targeted towards face recognition research. It differs from other datasets in that it contains face annotations for videos and video frames, unlike other datasets which only contain still images.  In \cite{vgg}, the authors released a dataset of over 2.6 million faces covering about 2,600 identities. However, this dataset contains much more label noise compared to \cite{celebfaces+} and \cite{casia}. 

Despite the availability of these datasets, there is still a need for more publicly available datasets to push the state-of-the-art forward. The datasets need to be more diverse in terms of head pose, occlusion, and quality of images. Also, there is a need to compare performance improvements with deep data (fewer subjects and more images per subject) against wide data (more subjects but fewer images per subject). 

The goal of this work is to introduce a new dataset \footnote {Available from ~\url{https://www.umdfaces.io}} which will facilitate the training of improved models for face recognition, head pose estimation, and keypoint localization (See figure \ref{fig:hyperface_output}). The new dataset has \umdnumim ~face annotations of \umdnumid ~subjects. Similar to \cite{casia}, our dataset is wide and may be used separately or to complement the CASIA dataset. We describe the data collection and annotation procedures and compare the quality of the dataset with some other available datasets. We will release this dataset publicly for use by the academic community. We provide bounding box annotations which have been verified by humans. Figure \ref{fig:five_people} shows a small sample of faces in the dataset for five subjects. We also provide the locations of fiducial keypoints, pose (roll,pitch and yaw) and gender information generated by the model presented in \cite{ultraface}. In addition to this, we also provide human verification of keypoint locations for \numkeypoints~images.

The rest of the paper is organized as follows. In section \ref{dat-col}, we describe the data collection procedure. We place this work in context with existing works in section \ref{rel-wor}. In section \ref{dat-stat}, we present the statistics of the dataset. We report the results of our baseline experiments in section \ref{exp} and in section \ref{disc}, we discuss the implications of the work and future extensions. 

\section{Data Collection}
\label{dat-col}
In this section we describe the data collection process and explain the semi-autonomous annotation procedure. We are releasing a total of \umdnumim ~images with face annotations spread over \umdnumid ~subjects. We provide bounding box annotations for faces which have been verified by human annotators. We are also releasing 3D pose information (roll, pitch, and yaw), twenty-one keypoint locations and their visibility, and the gender of the subject. These annotations have been generated using the All-in-one CNN model presented in \cite{ultraface}. 

\subsection{Downloading images}
Using the popular web-crawling tool, GoogleScraper \footnote {\url{https://github.com/NikolaiT/GoogleScraper}}, we searched for each subject on several major search engines (Yahoo, Yandex, Google, Bing) and generated a list of urls of images. We removed the duplicate urls and downloaded all the remaining images. 

\subsection{Face detection}
We used the face detection model proposed by Ranjan \etal to detect the faces in the downloaded images \cite{rajeev-btas}. Because we wanted a very high recall, we set a low threshold on the detection score. We kept all the face box proposals above this threshold for the next stage. 

%

\subsection{Cleaning the detected face boxes by humans}
Several bounding boxes obtained by the process discussed above do not contain any faces. Also, for each subject, there may be some detected face boxes which do not belong to that person. These cause noise in the dataset and need to be removed. We used Amazon Mechanical Turk (AMT) which is a widely used crowd-sourcing platform to get human annotations. These annotations are then used to remove extraneous faces.

For each subject, we showed six annotators batches of forty cropped face images. Out of these forty faces, thirty-five  were face detections which we suspected were images of the target subject but were not sure and five were added by us that we knew were not of the target individual. We knew the locations of these 5 `salt' images and used these to verify the quality of annotations by an annotator. We also displayed a reference image for that person which was selected manually by the authors. The annotators were asked to mark all the faces which did not belong to the subject in consideration. 

We evaluate the annotators by how often they marked the `salt' images that were presented to them. For example, if an annotator did 100 rounds of annotations and of the 500 `salt' images presented  he/she clicked on 496 of them, his/her vote was given a weight of 496/500.

To actually determine if a given image is of the target individual or not, we used the following robust algorithm which associated with every face a score between 0 and 1:

\begin{enumerate}
	\item Obtain the three highest vote weights and respective votes of all the annotators that had to decide on this face and call them $w_1$, $w_2$ and $w_3$, and their respective yes (1) - no (0) votes $v_1$, $v_2$ and $v_3$. For example $w_3$ is the vote weight of the highest scored annotator for this face, who voted for $v_3$.
	\item If $w_1 + w_2 > 0.8$, the final score of this face is $\sum_{i=1}^3 w_i v_i / \sum_{i=1}^3 w_i$
	\item If $w_3 > 0.6$, make the final score of this face $v_3$.
	\item Otherwise there is no reliable, robust answer for this face; try to annotate it again.
\end{enumerate}

This score has the following interpretation: closer to 0 means there is a robust consensus that the image is of the target individual and closer to 1 means that there is a robust consensus that it is an image not of the target individual.

After associating a score with every face we had, we selected the faces whose score was lower than 0.3 (after considering the quality and quantity trade-offs) and removed all other faces from our dataset.

The mechanism presented in this section allowed us to economically and accurately label all the faces we obtained.

In the next section we describe the method for generating other annotations. 

\subsection{Other annotations}
After obtaining the clean, human verified face box annotations, we used the all-in-one CNN model presented in \cite{ultraface} to obtain pose, keypoint locations, and gender annotations\footnote{We thank the authors of \cite{ultraface} for providing us the software for the all-in-one model.}. All-in-one CNN is the state-of-the-art method for keypoint localization and head pose estimation. 

We give a brief overview of this model.  

\textbf{All-In-One CNN:}
The all-in-one CNN for face analysis is a single multi-task model which performs face detection, landmarks localization, pose estimation, smile detection, gender classification, age estimation and face verification and recognition. For the task of face detection, the algorithm uses Selective Search  \cite{sel-search} to  generate region proposals from a given image and classifies them into face and non-face regions. Since we already have the cleaned detected face annotation, we pass it directly as an input to the algorithm. The all-in-one CNN uses this input to provide the facial landmark locations, gender information, and estimates the head pose (roll, pitch, yaw) in a single forward pass of the network. 

\begin{figure*}
	\begin{center}
		\includegraphics[width=0.85\linewidth,height=0.6\linewidth]{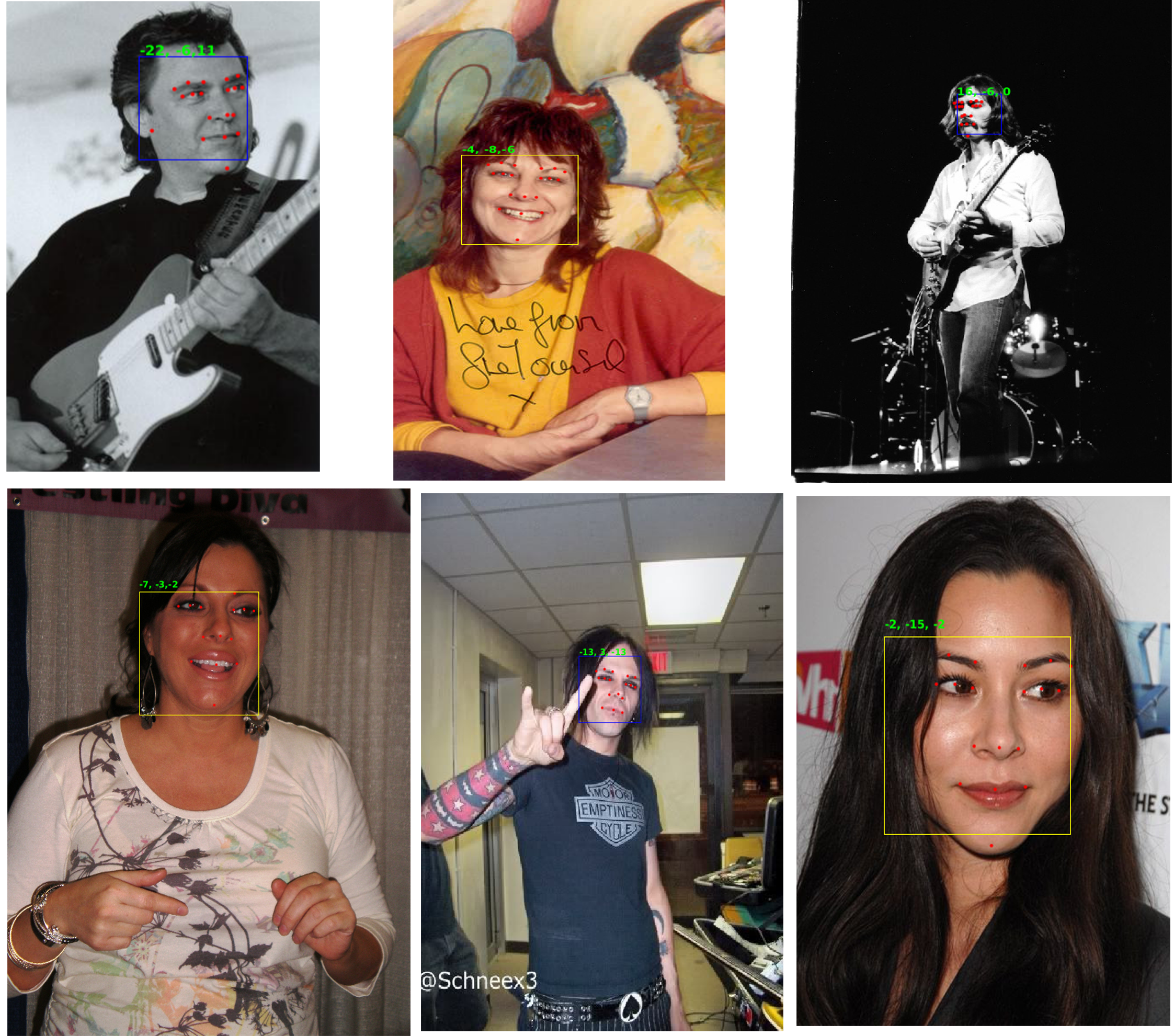}
	\end{center}
	\caption{Some examples with annotations generated by the all-in-one CNN \cite{ultraface}. The blue box indicates that the estimated gender is male and the yellow box means that the estimated gender is female. Red dots are the detected keypoints and the green text is the estimated head pose (yaw, roll, pitch).}
	\label{fig:hyperface_output}
\end{figure*}

Figure \ref{fig:hyperface_output} shows some examples of the annotations in our dataset generated by the all-in-one CNN algorithm. 

To verify the performance of the keypoints generated by the above model, we showed the generated annotations for \numkeypoints~images to humans and asked them to mark the images with incorrect keypoint annotations. We showed each face to two people on Amazon Mechanical Turk (AMT). As a mark of the quality of the keypoints, we found that for about 28,084 images out of the 115,000 shown did both the annotators say that the keypoints are incorrectly located. We will publicly release this data collected from AMT. This will enable researchers working on face recognition and analysis problems to improve performance.

\subsection{Final cleaning of the dataset}
We noticed that even after getting human annotations, the dataset still had some noisy face bounding boxes. For some individuals there were some boxes that belonged to someone else or were not faces at all. Since we wanted to provide the cleanest dataset that we could, we removed these noisy boxes. Here we present the approach that was taken to remove them.

The face verification problem has been studied for a very long time now. One-to-one face verification is the most commonly studied problem in verification \cite{lfw, ytf}. Several algorithms are achieving better-than-human performance on the LFW dataset \cite{lfw} which was an early benchmark for face verification \cite{facenet, deepface, surpassing, joint, deeply, deepid_3}.

\begin{figure}
	\begin{center}
		\includegraphics[width=0.7\linewidth]{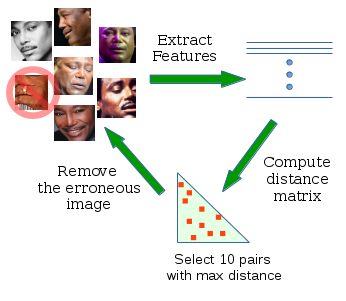}
	\end{center}
	\caption{Overview of the strategy for final cleaning of the dataset.}
	\label{fig:fin_clean}
\end{figure}

We used the verification model proposed in \cite{swami} to remove the noise. The network trained in \cite{swami} is targeted towards IJB-A \cite{ijba} which is a much tougher dataset than LFW. For each subject , we extracted the fc7 layer features and calculate the cosine distance ($1-cos(\theta)$), where $\theta$ is the angle between the two feature vectors) between each pair of faces for that subject. We found the ten pairs with the maximum distance between them and sum these ten distances. We observed that if this sum is below a certain threshold (ten in our tests), then all the pairs are actually images of the same person. However, if the sum is above the threshold, then most of the times there is at least one noisy face box in the data for that subject. So, if the sum of distances was above the threshold, we found the face image that occurs in the maximum number of pairs out of the ten pairs selected and removed that image from the dataset. If more than one image occurred the maximum number of times, then we removed the one which contributes the most to the sum. We again calculate the similarity matrix and repeat the process till the sum of the ten pairs goes below the threshold. Figure \ref{fig:fin_clean} summarizes this approach.

If the above procedure led to the removal of more than five images for a subject then we removed that subject id. Using this process we removed \umremovedim ~images and \umremovedsub ~subject identities from the dataset. Finally, our dataset has \umdnumim ~face annotations spread over \umdnumid ~subject identities.

We divide the dataset into non-overlapping `train' and `test' parts. We will release this division and the testing protocol to be used by researchers as a tougher evaluation metric than some existing metrics. In section \ref{exp:recog}, we use the `train' set to train a deep network for verification and compare its performance against a network trained on CASIA WebFace \cite{casia} and an off-the-shelf network \cite{vgg}. We evaluate the performance of all three networks on the `test' set of our dataset. We show that the network trained on the UMDFaces dataset achieves the best verification performance of the three. Our model is a benchmark on the `test' set of our dataset.

\section{Related Works}
\label{rel-wor}
There is a dearth of publicly available high quality large face datasets. An overview of the most widely used publicly available face datasets is presented in table \ref{tab:rel-work}. 

\begin{table*}
	\begin{center}
		\begin{tabular}{|c|c|c|c|}
			\hline
			\textbf{Dataset} & \textbf{Number of subjects} & \textbf{Number of images} & \textbf{Annotation Properties} \\
			\hline\hline
			VGG Face \cite{vgg} & 2,622 & 2.6 million & Bounding boxes and coarse pose \\ \hline
			CASIA WebFace \cite{casia} & 10,575 & 494,414 & - \\ \hline
			CelebA \cite{celebfaces+,celebA} & 10,177 & 202,599 & 5 landmarks, 40 binary attributes \\ \hline
			FDDB \cite{fddb} & - & 2,845 (5,171 faces) & Bounding boxes \\ \hline
			WIDER FACE \cite{wider} & - & 32,203 (about 400,000 faces) & Bounding boxes and event category \\ \hline
			IJB-A \cite{ijba} & 500 & 24,327 (49,759 faces) & Face boxes, and eye and nose\\ &&& locations \\ \hline
			LFW \cite{lfwsurvey} & 5,749 & 13,233 & Several attribute annotations \\ \hline
			AFLW \cite{aflw} & - & 25,993 & Bounding boxes and 21 keypoint\\ &&& locations \\ \hline
			YTF \cite{ytf} & 1,595 & 3,425 videos & - \\ \hline
			MSCeleb \cite{msceleb, msceleb1} & 100,000 (training set) & 10 million & - \\ \hline
			MegaFace \cite{megaface2} & 672,057 & 4.7 million & - \\ \hline
			\textbf{Ours} & \umdnumid & \umdnumim & Bounding boxes, 21 keypoints, \\ &&& gender and 3D pose \\ \hline
			\hline
		\end{tabular}
	\end{center}
	\caption{Recent face detection and recognition datasets.}
	\label{tab:rel-work}
\end{table*}

There are basically two problems that face researchers focus on. These are (1) face detection (including keypoint location estimation), and (2) face recognition. Our dataset has annotations for identity, face bounding boxes, head pose, and keypoint locations. The dataset can benefit researchers working on face recognition or keypoint localization problems. We do not provide bounding boxes for all the faces in an image, but just for one subject. This means that our dataset is not suitable for training face detection models. The scale variation in our dataset is also less than some other datasets which are specifically targeted at the detection problem. Now we discuss the available datasets separately based on the problem they are targeted at.

\textbf{Detection:} The most popular datasets used for face detection are WIDER FACE \cite{wider}, FDDB \cite{fddb}, and IJB-A \cite{ijba}. The WIDER FACE dataset contains annotations for 393,703 faces spread over 32,203 images. The annotations include bounding box for the face, pose (typical/atypical), and occlusion level (partial/heavy). FDDB has been driving a lot of progress in face detection in recent years. It has annotations for 5,171 faces in 2,845 images. For each face in the dataset, FDDB provides the bounding ellipse. However, FDDB does not contain any other annotations like pose. The IJB-A dataset was introduced targeting both face detection and recognition. It contains 49,759 face annotations over 24,327 images. The dataset contains both still images and video frames. IJB-A also does not contain any pose or occlusion annotations.

AFLW \cite{aflw} is the dataset closest to our dataset in terms of the information provided. There are 25,993 labeled images in the dataset. AFLW provides annotations for locations of 21 keypoints on the face. It also provides gender annotation and coarse pose information. 

Our dataset is about 15 times larger than AFLW. We provide the face box annotations which have been verified by humans. We also provide fine-grained pose annotations and keypoint location annotations generated using the all-in-one CNN \cite{ultraface} method. The pose and keypoint annotations haven't been generated using humans as annotators. However, in section \ref{dat-stat} we analyze the accuracy of these annotations. This dataset can be used for building keypoint localization and head pose estimation models. We compare a model trained on our dataset with some recent models trained on AFLW in terms of keypoint localization accuracy in section \ref{exp}.

\textbf{Recognition:} There has been a lot of attention to face recognition for a long time now. Face recognition itself is composed of two problems: face identification and face verification. With the advent of high capacity deep convolutional networks, there is a need for larger and more varied datasets. The largest datasets that are targeted at recognition are the ones used by Google \cite{facenet} and Facebook \cite{deepface}. But these are not publicly available to researchers.

However, recently, Microsoft publicly released the largest dataset targeted at face recognition \cite{msceleb}. It has about 10 million images of 100,000 celebrities. However, the authors of \cite{msceleb} did not remove the wrong images from the dataset because of the scale of the dataset. Since this dataset is so new, it remains to be seen whether models which are robust to such large amounts of noise could be developed. Another large scale dataset targeted at recognition is the VGG Face dataset \cite{vgg}. It has 2.6 million images of 2,622 people. But, the earlier version of this dataset had not been completely curated by human annotators and contained label noise. The authors later released the details about curation of the dataset and finally there are just about 800,000 images that are in the curated dataset. This number makes it among the largest face datasets publicly available. The dataset is very deep in the sense that it contains several hundreds of images per person. On the other hand, our dataset is much wider (more subjects and fewer images per subject). An interesting question to be explored is how a deep dataset compares with a wide dataset as a training set. The authors of \cite{vgg} also provide a pose annotation (frontal/profile) for each face. But the dataset is not very diverse in terms of pose and contains 95\% frontal images and just 5\% non-frontal faces. 

The recently released Megaface challenge \cite{megaface1} might be the most difficult recognition (identification) benchmark currently. Megaface dataset is a collection of 1 million images belonging to 1 million people. This dataset is not meant to be used as training or testing dataset but as a set of distractors in the gallery image set. Megaface challenge uses the Facescrub \cite{facescrub} dataset as the query set. The MegaFace challenge also lead to the creation of another large dataset which has over 4.7 million images of over 670,000 subjects \cite{megaface2}.

The two datasets which are closest to our work are CASIA WebFace \cite{casia} and CelebFaces+ \cite{celebfaces+} datasets. The CASIA WebFace dataset contains 494,414 images of 10,575 people. This dataset does not provide any bounding boxes for faces or any other annotations.  Celebfaces+ contains 10,177 subjects and 202,599 images. CelebA \cite{celebA} added five landmark locations and forty binary attributes to the CelebFaces+ dataset. 

YouTube Faces (YTF) is another dataset that is targeted towards face recognition. However, it differs from all other datasets because it is geared towards face recognition from videos. It has 3,425 videos of 1,595 subjects. The subject identities in YTF are a subset of the subject identities in LFW.  

\section{Dataset Statistics}
\label{dat-stat}
In this section, we first discuss the performance of the all-in-one CNN model used to generate the keypoints and pose annotations in our dataset. Then we evaluate some statistics of the proposed dataset and compare them with those of similar datasets. In section \ref{exp:key}, we will also demonstrate that using these annotations as training data, we can get better performance for a keypoint location detector than when just using AFLW as the training set.

The authors of \cite{ultraface} compare the performance of their keypoint detector with the performance of other algorithms and report state-of-the-art results on AFLW (Table II in \cite{ultraface}). Our hypothesis is that the keypoints predicted using the all-in-one CNN model \cite{ultraface} for our dataset, we can create a better keypoint detection training dataset than AFLW \cite{aflw}. We verify this in section \ref{exp:key} where we train a bare-bones network using our dataset as the training data for keypoint localization.

\begin{figure}[t]
	\begin{center}
		\includegraphics[width=\linewidth,height=0.7\linewidth]{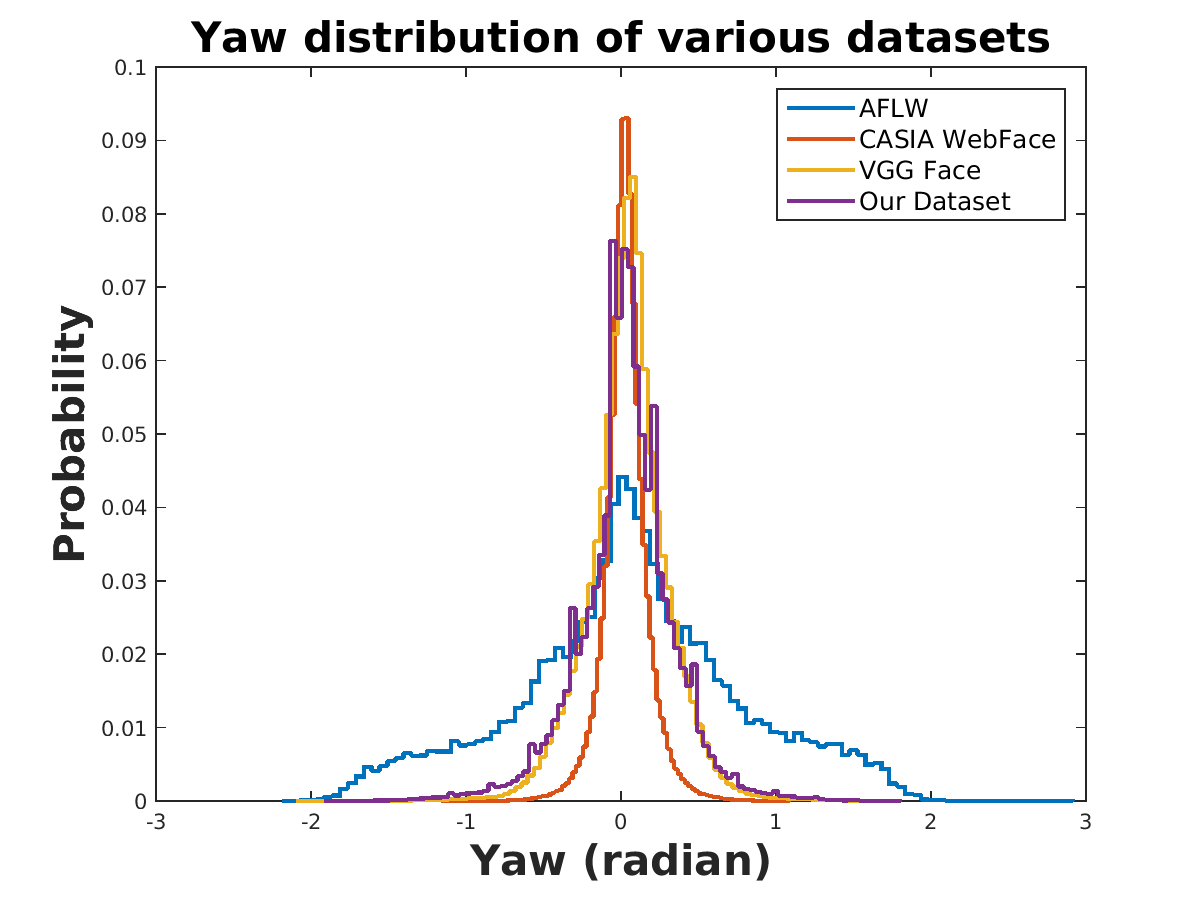}
	\end{center}
	\caption{Histogram of the yaw angles of the faces in four datasets. The yaws in our dataset are more spread-out than the yaws in CASIA WebFace \cite{casia} and almost the same as VGG Face \cite{vgg}. AFLW \cite{aflw} has a much wider distribution but it is very small compared to the other datasets and does not provide any identity information.}
	\label{fig:yaw_hist}
\end{figure}

Figure \ref{fig:yaw_hist} shows the distribution of the yaw angles of the head in four datasets. We note that the distribution of the yaw angles in our dataset is much wider than the distribution in CASIA WebFace \cite{casia} which is a dataset similar in size to ours. Also note that, the distribution is almost the same as in VGG Face \cite{vgg} even though it is a deeper (more images per subject) dataset. An interesting question that can be explored in the future is whether the depth in VGG provides any advantages for training recognition models. 

\begin{figure}[t]
	\begin{center}
		\includegraphics[width=\linewidth,height=0.6\linewidth]{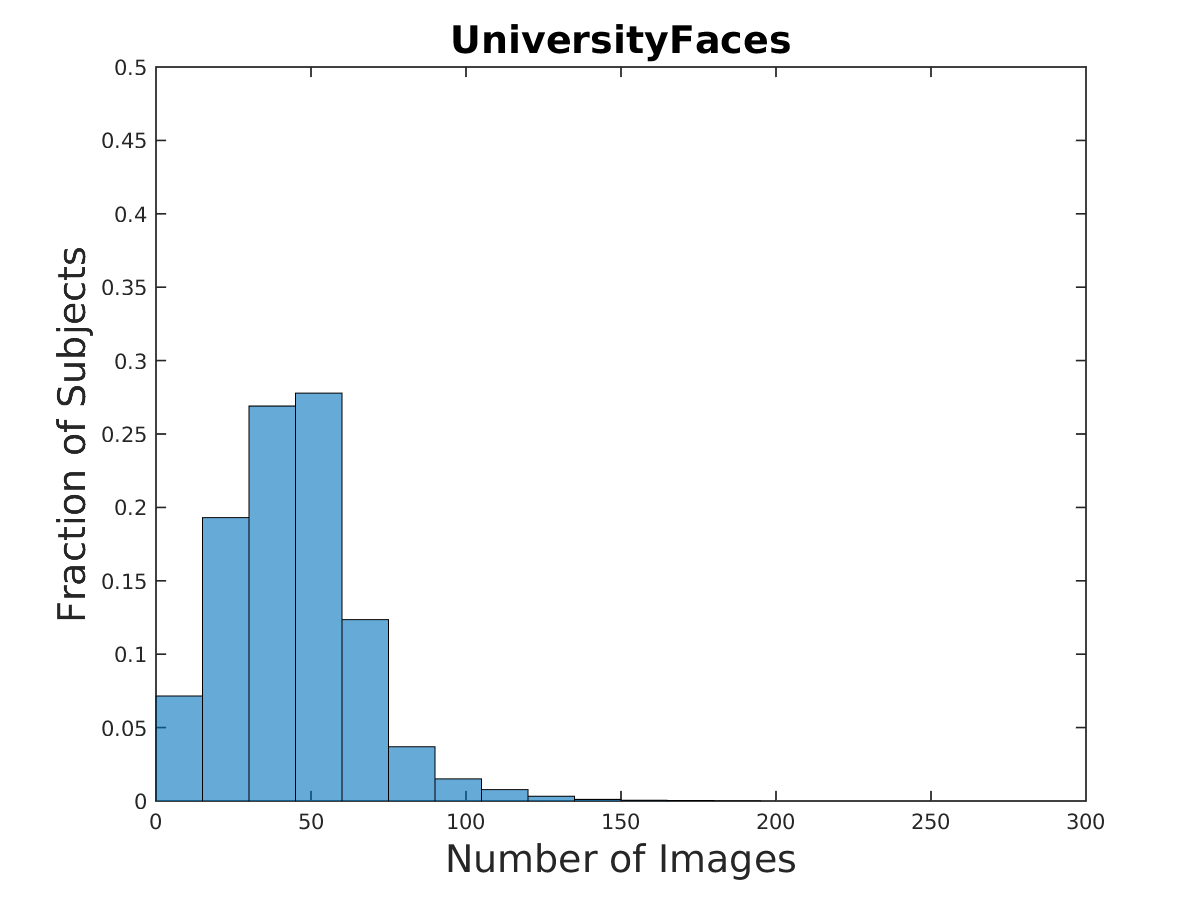}
	\end{center}
	\caption{Histogram of the number of face annotations per subject in our dataset.}
	\label{fig:num_hist_umd}
\end{figure}

\begin{figure}[t]
	\begin{center}
		\includegraphics[width=\linewidth,height=0.6\linewidth]{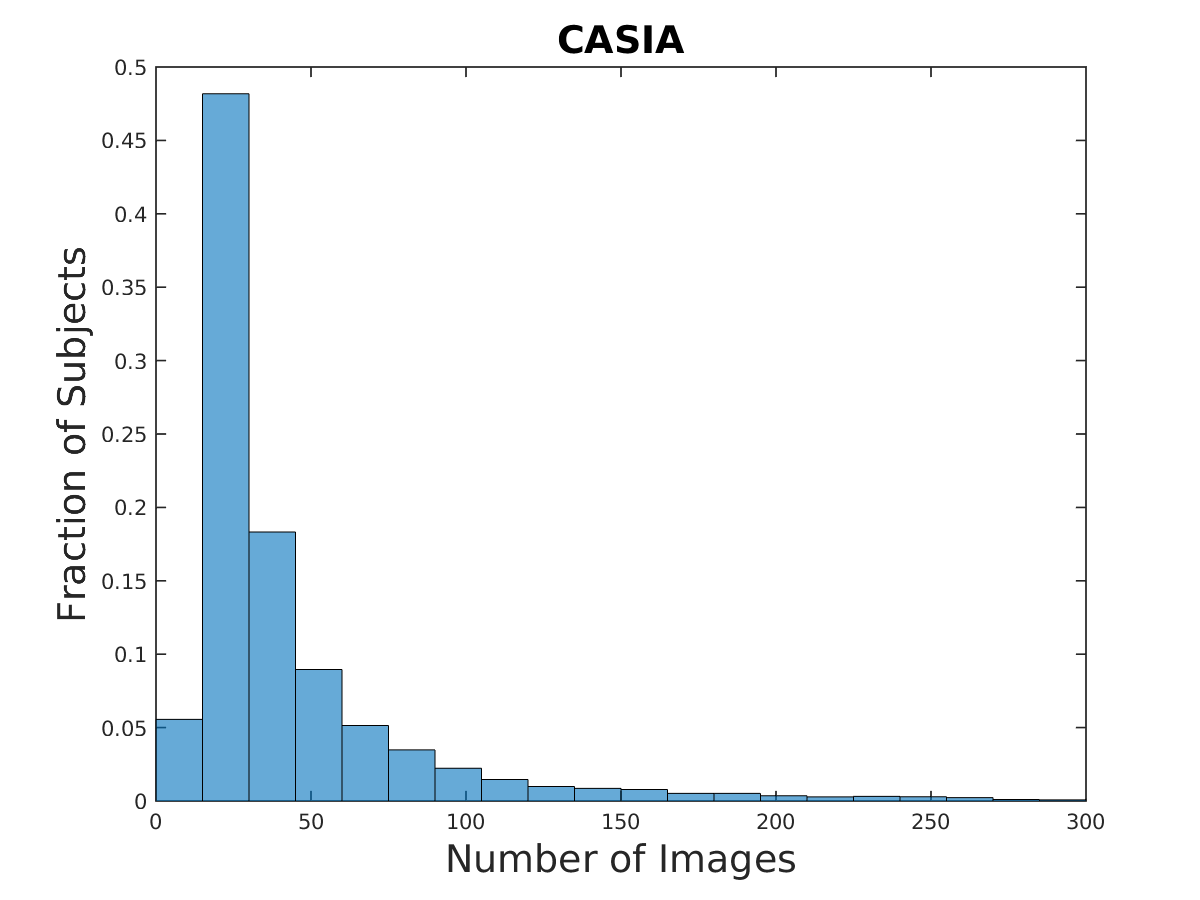}
	\end{center}
	\caption{Histogram of the number of face annotations per subject in CASIA WebFace \cite{casia}.}
	\label{fig:num_hist_casia}
\end{figure}

Figure \ref{fig:num_hist_umd} shows the distribution of the number of face annotations per subject in our dataset. We note that this distribution is relatively uniform around the 50 images per subject mark and it is not skewed towards very few subjects containing most face annotations as is the case for CASIA WebFace dataset \cite{casia} (figure \ref{fig:num_hist_casia}).

\section{Experiments}
\label{exp}
We evaluate the quality of our dataset by performing some baseline experiments. First, we show that a deep network trained on our dataset performs better than a similar network trained on CASIA WebFace \cite{casia} and an off-the-shelf VGG Face network \cite{vgg}. Then we show the quality of our keypoints by training a deep network on the provided keypoints and achieving near state-of-the-art performance on keypoint-location prediction.

\subsection{Face Verification}
\label{exp:recog}
We train a recognition network based on the Alexnet architecture \cite{alexnet} on a subset of our dataset which we call the `train' set and another network on the CASIA WebFace dataset \cite{casia}. We use these networks and an off-the shelf network trained on VGGFace dataset \cite{vgg} to compare face verification performance on a disjoint subset of our dataset which we call the `test' set. The authors in \cite{vgg} mention that aligning faces during training is not necessary and aligning the faces while testing improves performance. We use faces aligned using keypoints from \cite{ultraface} while testing. Now, we briefly describe our test protocol.

\subsubsection{Test Protocol}
While we acquired and curated UMDFaces to be primarily a training  dataset, we also developed a testing protocol on top of it, specifically on top of a subset of it. We define a large verification protocol, that contains three tracks:
\begin{itemize}
	\item \textbf{Small pose variation (Easy):} Absolute value of the yaw difference $\Delta \in [0, 5)$ (all angles expressed in degrees)
	\item \textbf{Medium pose variation (Moderate):} Absolute value of the yaw difference $\Delta \in [5, 20)$ (all angles expressed in degrees)
	\item \textbf{Large pose variation (Difficult):} Absolute value of the yaw difference $\Delta \in [20, \infty)$ (all angles expressed in degrees)
\end{itemize}

Each of the three tracks has a total of 50,000 positive (same individual) pairs and 50,000 negative (different individual) pairs. The benefit of selecting a large number of total pairs of images for evaluation is that it allows for a comparison of the performance at very low false accept rates.

We envision that researchers will evaluate on the UniverstiyFaces protocol and that evaluating on UMDFaces would show how robust different methods are to a more difficult selection of faces.

We will release the testing protocol along with the UMDFaces dataset.

To generate the protocol, we used \numtestid~random subjects (\numtest~faces) from the UMDFaces dataset. For each face of each individual we computed the yaw using the method described in \cite{ultraface}. For each of the three tracks we randomly selected 50,000 intra-personal pairs that satisfied the absolute value of the yaw difference for the track and 50,000 extra-personal pairs that satisfied the absolute value of the yaw difference for the track. 

We use the method used in \cite{swami} for evaluation. After training a network, we pass each face image in a test set through the network and extract the feature vector from the last fully connected layer before the classification layer. We use these feature vectors for a pair of images to compute similarity between two faces using the cosine similarity metric. We use ROC curves as our performance metric. 

\begin{figure}[t]
	\begin{center}
		\includegraphics[width=\linewidth,height=0.8\linewidth]{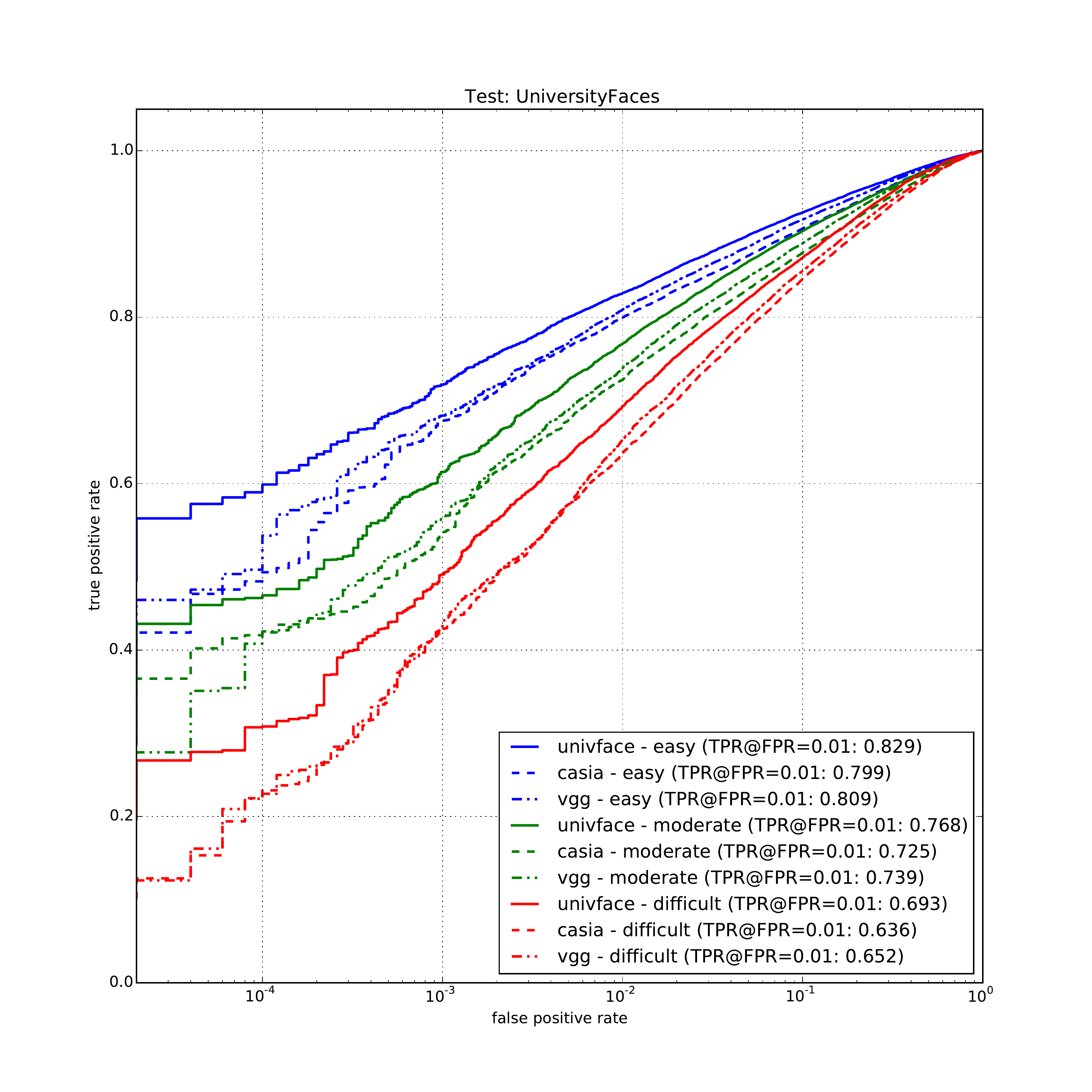}
	\end{center}
	\caption{Performance evaluation on the `test' set of our dataset. The three colours represent easy (blue), moderate (green), and difficult (red) test cases. `Easy' represents the case where the difference in yaw of the two images is less than 5 degrees. `Moderate' represents a yaw difference between 5 and 20 degrees and `difficult' means that the yaw difference is more than 20 degrees.}
	\label{fig:umd_batch3}
\end{figure}

Figure \ref{fig:umd_batch3} shows the performance of the three networks on the `test' set of our dataset. We see that the network trained on our dataset performs better than both the network trained on CASIA WebFace and the off-the-shelf network trained on VGGFace. The difference is particularly apparent at low false acceptance rates where the network trained on UMDFaces dataset significantly outperforms the other two models (for example see $FPR=10^{-4}$ in figure \ref{fig:umd_batch3}). 

\begin{figure}[t]
	\begin{center}
		\includegraphics[width=\linewidth,height=0.65\linewidth]{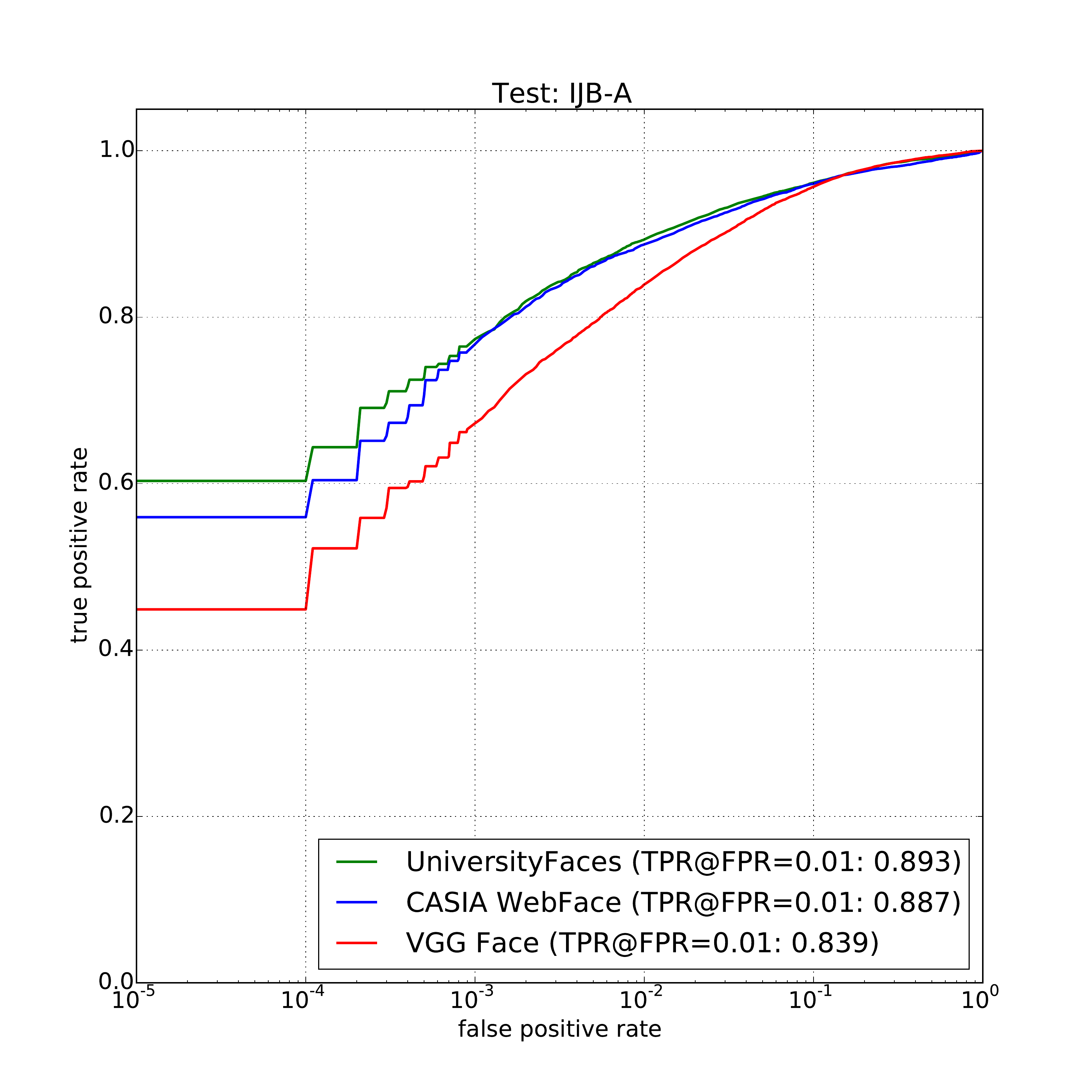}
	\end{center}
	\caption{Performance on the IJB-A evaluation protocol \cite{ijba}.}
	\label{fig:ijba}
\end{figure}

We also train another model on our complete dataset of \umdnumid~images and evaluate it on the IJB-A evaluation protocol \cite{ijba}. Figure \ref{fig:ijba} shows the comparison of our model with the previously mentioned models trained on CASIA WebFace and  VGGFace. Again, our model performs better than the other two networks across the board and particularly for low false acceptance rates.

We observe that the protocol used here is a tougher evaluation criterion than existing ones like LFW \cite{lfw} and IJB-A \cite{ijba}. Using this protocol for evaluating the performance of deep networks will help push the face recognition and verification research forward.
%

\subsection{Keypoint Detection}
\label{exp:key}
We train a simple deep convolutional neural network for keypoint localization using all of the released dataset as the training set and compare the accuracy of the model with the accuracy of some recent models trained using the AFLW dataset \cite{aflw}. We evaluate the performance on the ALFW test dataset and the AFW \cite{afw} dataset. We demonstrate that just this simple network trained on our dataset is able to perform comparably or even better than several recent systems which are much more complex and use several tricks to achieve good performance. 

\begin{figure}[t]
	\begin{center}
		\includegraphics[width=\linewidth,height=0.4\linewidth]{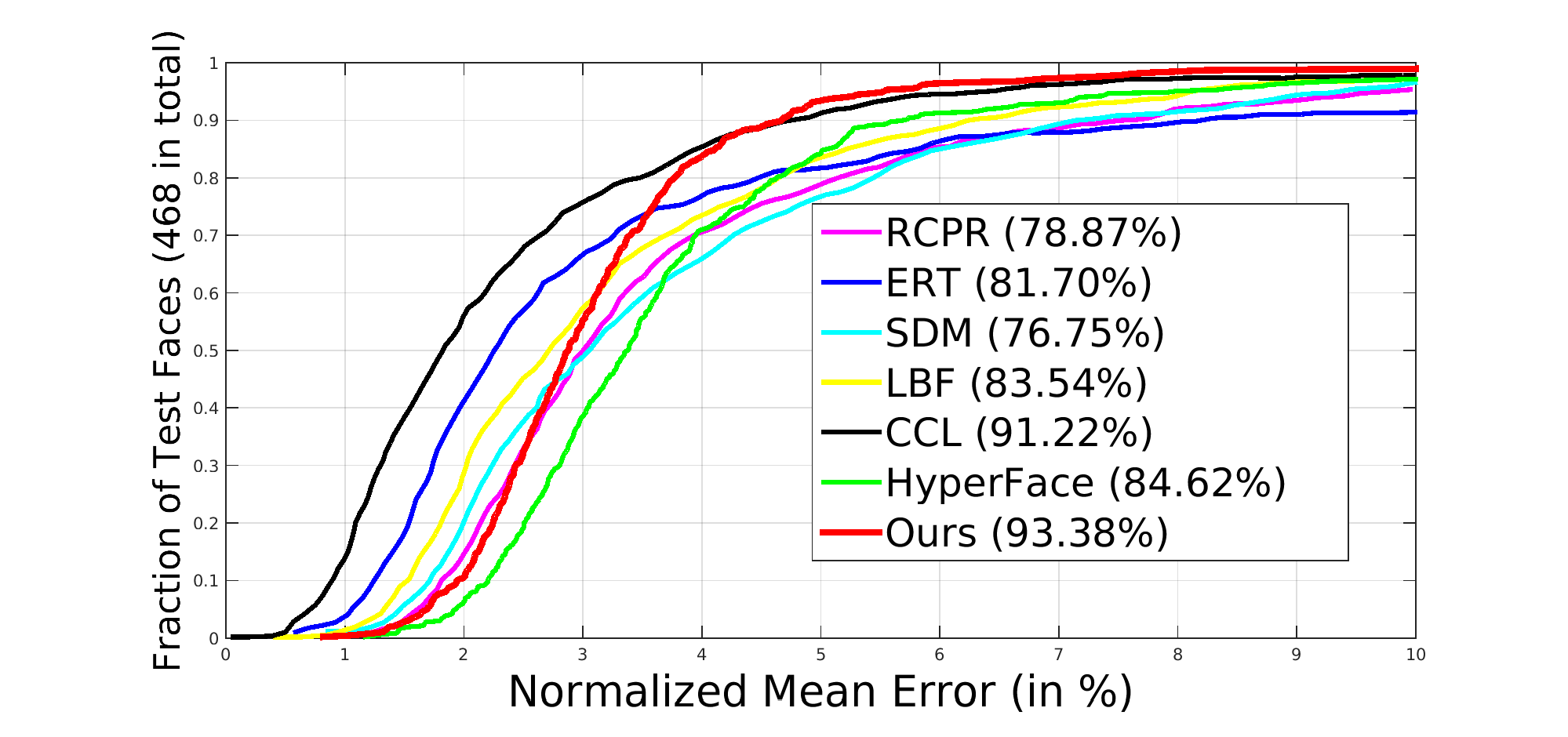}
	\end{center}
	\caption{Performance evaluation on AFW dataset (6 points) for landmarks localization task. The numbers in the legend are the percentage of test faces with NME less than $5\%$.}
	\label{fig:exp_plot}
\end{figure}

We used the commonly used VGG-Face \cite{vgg} architecture and changed the final layer to predict the keypoints. We trained the network on our dataset till it converged. Figure \ref{fig:exp_plot} shows the performance of recent keypoint localization methods on the AFW dataset \cite{afw}. We note that our model out-performs all the recently published methods at a normalized mean error of 5$\%$. In table \ref{tab:exp-tab}, we compare the performance of our model on the AFLW keypoint localization test dataset. Our model performs comparably or better than all recently published methods. We will release the network weights publicly. 

\begin{table}
	\label{tab:exp-tab}
	\begin{center}
		\resizebox{\linewidth}{!}{%
		\begin{tabular}{|c|c|c|c|c|c|}
			\hline
			& \multicolumn{5}{c|}{AFLW Dataset (21 points)} \\ \hline
			\textbf{Method}             & {[}0, 30{]}   & {[}30, 60{]}  & {[}60, 90{]}  & mean          & std           \\ \hline
			RCPR~\cite{burgos2013robust}              & 5.43          & 6.58          & 11.53         & 7.85          & 3.24          \\ \hline
			ESR~\cite{DBLP:journals/ijcv/CaoWWS14}                & 5.66          & 7.12          & 11.94         & 8.24          & 3.29          \\ \hline
			SDM~\cite{XiongD13}               & 4.75          & 5.55          & 9.34          & 6.55          & 2.45          \\ \hline
			3DDFA~\cite{zhu2015face}              & 5.00          & 5.06          & 6.74          & 5.60          & 0.99          \\ \hline
			3DDFA~\cite{3ddfa}+SDM~\cite{sdm}          & 4.75          & 4.83          & 6.38          & 5.32          & 0.92          \\ \hline
			HyperFace~\cite{hyperface} 		   & 3.93          & 4.14          & 4.71          & 4.26 			& 0.41          \\ \hline
			\textbf{Ours} & 4.39 & 4.81 & 5.50          & 4.90 	& 0.56          \\ \hline
		\end{tabular}
	}
	\end{center}
	\caption{The NME(\%) of face alignment results on AFLW test set for various poses (frontal ({[}0-30{]}) to profile ({[}60-90{]})).}
\end{table}

This experiment highlights the quality of the data and provides baseline results for fiducial landmark localization. By training just a bare-bones network on our dataset we are able to achieve good performance. This shows that this dataset will be very useful to researchers working in this area for obtaining improved models.  

\section{Discussion}
\label{disc}
In this work we release a new dataset for face recognition and verification. We provide the identity, face bounding boxes, twenty-one keypoint locations, 3D pose, and gender information. Our dataset provides much more variation in pose than the popular CASIA WebFace \cite{casia} dataset. This will help researchers achieve improved performance in face recognition. We release a new test protocol for face verification which is tougher than the most commonly used protocols. We show the importance of our dataset by comparing deep verification networks trained on various similarly sized datasets. We also demonstrate the quality of the automatically generated keypoint locations by training a simple CNN and comparing its performance with recent algorithms which are very complex. We believe that using the presented dataset, these complex models can achieve even better performance. Additionally, we also verify the quality of the keypoint annotations for part of the data.
%


\section*{Acknowledgments}
This research is based upon work supported by the Office of the
Director of National Intelligence (ODNI), Intelligence Advanced Research Projects Activity (IARPA), via IARPA R\&D Contract No.
2014-14071600012. The views and conclusions contained herein are
those of the authors and should not be interpreted as necessarily
representing the official policies or endorsements, either expressed
or implied, of the ODNI, IARPA, or the U.S. Government. The U.S.
Government is authorized to reproduce and distribute reprints for
Governmental purposes notwithstanding any copyright annotation
thereon.

{\small
\bibliographystyle{ieee}
\bibliography{egbib,UltraFace}
}

\end{document}